\documentclass[letterpaper]{article} 
\usepackage{aaai24}  
\usepackage{times}  
\usepackage{helvet}  
\usepackage{courier}  
\usepackage[hyphens]{url}  
\usepackage{graphicx} 
\usepackage{natbib}  
\usepackage{caption} 
\usepackage{algorithm}
\usepackage{algorithmic}

\usepackage{amssymb}
\usepackage{amsmath}
\usepackage{booktabs}
\usepackage{multirow}

\def\eg{\emph{e.g.}} 
\def\ie{\emph{i.e.}}

\def\wrt{\emph{w.r.t.}} 

\def\CircleArrowright{\ensuremath{
  \rotatebox[origin=c]{310}{$\circlearrowright$}}}
\newcommand{\rvlnbert}{VLN$\protect\CircleArrowright$BERT}

\title{WebVLN: Vision-and-Language Navigation on Websites}
\author{
    Qi Chen\equalcontrib, ~Dileepa Pitawela\equalcontrib, ~Chongyang Zhao\equalcontrib, ~Gengze Zhou, ~Hsiang-Ting Chen, ~Qi Wu\thanks{Corresponding author.}
}
\affiliations{
    Australian Institute for Machine Learning, The University of Adelaide\\
    \{qi.chen04,~dileepa.pitawela,~chongyang.zhao,~gengze.zhou,~tim.chen,~qi.wu01\}@adelaide.edu.au
}

\usepackage{bibentry}

\begin{document}

\maketitle

\begin{abstract}
Vision-and-Language Navigation (VLN) task aims to enable AI agents to accurately understand and follow natural language instructions to navigate through real-world environments, ultimately reaching specific target locations. 
We recognise a promising opportunity to extend VLN to a comparable navigation task that holds substantial significance in our daily lives, albeit within the virtual realm: navigating websites on the Internet.
This paper proposes a new task named Vision-and-Language Navigation on Websites (WebVLN), where we use question-based instructions to train an agent, emulating how users naturally browse websites.
Unlike the existing VLN task that only pays attention to vision and instruction (language), the WebVLN agent further considers underlying web-specific content like HTML, which could not be seen on the rendered web pages yet contains rich visual and textual information.
Toward this goal, we contribute a dataset, WebVLN-v1, and introduce a novel approach called Website-aware VLN Network (WebVLN-Net), which is built upon the foundation of state-of-the-art VLN techniques.
Experimental results show that WebVLN-Net outperforms current VLN and web-related navigation methods.
We believe that the introduction of the new WebVLN task and its dataset will establish a new dimension within the VLN domain and contribute to the broader vision-and-language research community. 
Code is available at: https://github.com/WebVLN/WebVLN.
\end{abstract}

\section{Introduction}

Vision-and-Language Navigation (VLN)~\cite{anderson2018vision} aims to seamlessly integrate visual perception and action with language understanding, to enable AI agents to navigate and interact effectively within real-world environments.
Interestingly, the resemblance can be found in the virtual online environment, where users might rely on AI agents to assist them in gathering information about certain products even when they can only offer broad and vague instructions such as \textit{``how much does a pair of grey and orange striped socks cost''}.
This extends beyond the boundaries of traditional VLN tasks, including not only vision and instruction (language) but also incorporating the abundant information embedded within webpage like HTML. With this consideration, we introduce an extended VLN task, denoted as Vision-and-Language Navigation on Websites (WebVLN).

\begin{figure}
    \centering
    \includegraphics[width=0.99\linewidth]{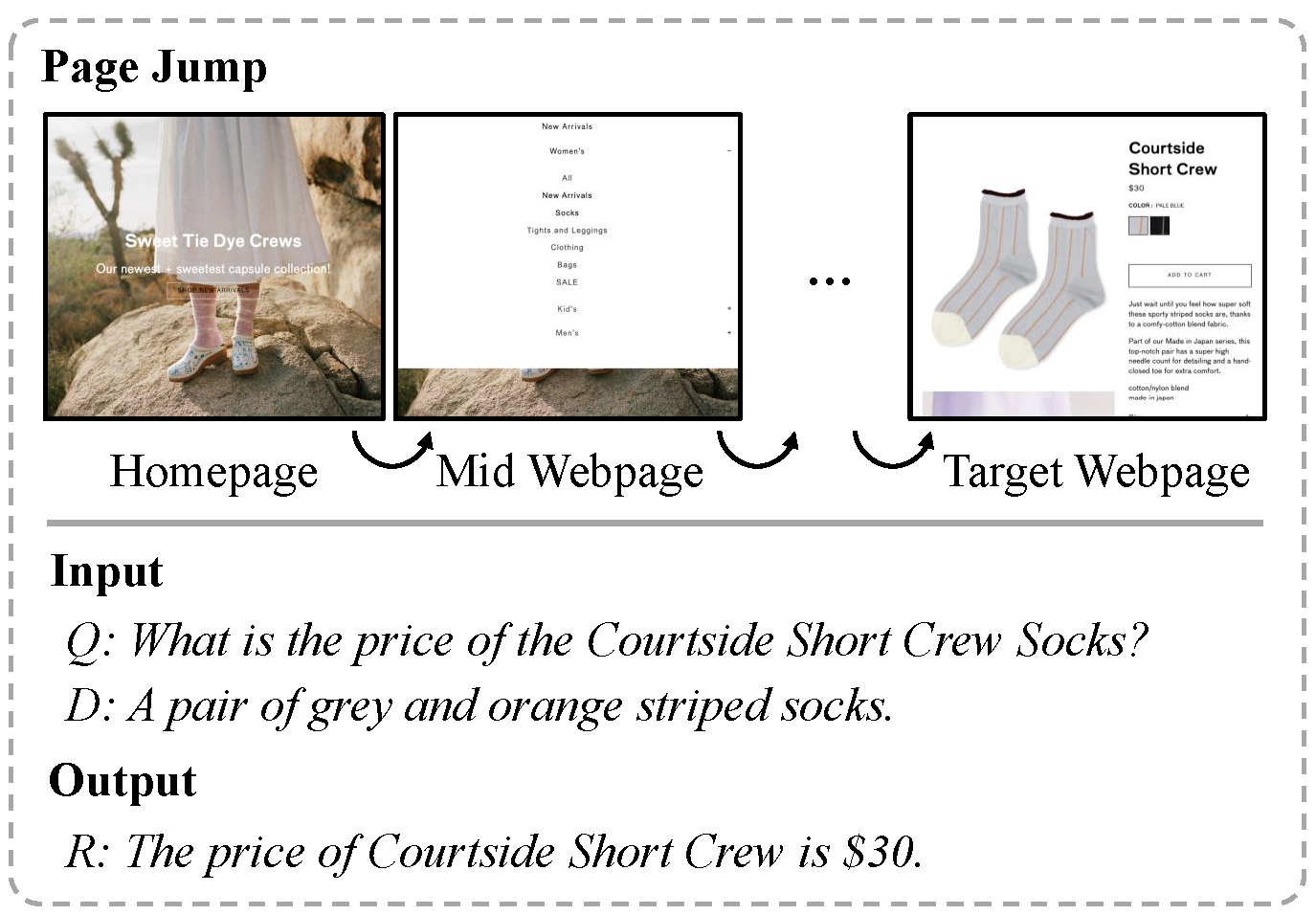}
    \caption{An example of the WebVLN task. An agent is initiated on the homepage of a website and asked a question $Q$ with an auxiliary description $D$. To respond, the agent is required to intelligently navigate and explore the website, gather information through observation, and finally provide an accurate response/answer $R$ in a free-form sentence.}
    \label{fig:example}
\end{figure}

Figure~\ref{fig:example} shows an example of the WebVLN task. In this scenario, an agent starts its journey from a website's front page, presented with a question $Q$ accompanied by an auxiliary description $D$. 
The agent emulates genuine user behaviour and navigates through the website. It processes the current view of the webpage and engages in common web browsing activities such as reading the images and text, and clicking on the links to navigate to the next pages.
The agent's objective is to efficiently traverse the website and reach a target webpage, which contains the necessary information to answer the question $Q$ and produce an accurate response $R$ to the question.

The new task poses several new challenges. \textbf{First}, the choices available to an AI agent navigating a website are substantially more than those in traditional discrete VLN scenarios, which are confined to adjacent navigable viewpoints in physical environments.
While in WebVLN, the range of choices for each observed webpage is significantly broader as it contains a vast array of content, features, links, and interactive elements. Each webpage offers multiple avenues for navigation, such as clicking on various links, buttons, and dropdown menus.
\textbf{Second}, due to the intrinsic diversity of available choices on each webpage, WebVLN constructs a more intricate and complex navigation graph than traditional VLN, making it nearly impossible to explore all the content on websites by a naive heuristic trial-and-error scheme.
Thus, in the WebVLN task, an ideal method should seek to maximise accurate choices while minimising the need for exploration by leveraging varied information available within the webpage.

Due to the lack of an off-the-shelf dataset for WebVLN, we have collected a new WebVLN-v1 dataset to facilitate research in this field. It comprises $8,990$ records/paths with $14,825$ QA pairs derived from three different shopping websites (aliased as SA, HB and ES).
Differing from other VLN datasets~\cite{anderson2018vision,qi2020reverie} that only consider the visual inputs of the environment, our WebVLN-v1 incorporates both visual and textual contents extracted from the websites.
Furthermore, in comparison to other web-related datasets, such as web navigation~\cite{liu2018reinforcement,xu2021grounding,mazumder2020flin,yao2022webshop} and web QA~\cite{chang2022webqa,hsiao2022screenqa}, our WebVLN-v1 seamlessly integrates both navigation and QA environments with question-based instructions, resulting in a unified benchmark.

To tackle the challenging WebVLN task, we propose a new method called Website-aware Vision-and-Language Navigation Network (WebVLN-Net) based on the widely used VLN framework--\rvlnbert~\cite{hong2021vln}. Besides the visual input (screenshot) and instruction (question \& description), WebVLN-Net considers the underlying HTML of each webpage and extracts elements such as clickable buttons. Upon reaching a ``stop'' token, our model initiates answering the question using information from both click history and the current ``stop'' webpage.
The evaluation of the model performance is based on the metrics from both VLN and VQA domains. Specifically, for VLN, we consider success rate (SR), oracle success rate (OSR), success rate weighted by path length (SPL), and Trajectory Length (TL), while adopting Wu-Palmer Similarity (WUPS)~\cite{wu1994verb} for VQA evaluation due to the open-end setting, \ie, generating a free-form sentence as an answer.

In summary, our contributions include:
\begin{itemize}
    \item A new task - Vision-and-Language Navigation on Websites (WebVLN), where the agent emulates human web browsing behaviours and navigates to a specified target webpage based on the input question and its auxiliary description, subsequently answering the question using information extracted from the target webpage.
    \item A new WebVLN-v1 dataset, consisting of $8,990$ records/paths, and $14,825$ question-answer (QA) pairs derived from three different websites, covering both navigation and QA on the web environments. 
    \item A new method, named Website-aware VLN Network (WebVLN-Net), which not only considers the visual input (screenshot) and linguistic instruction but also uses web-specific content (\ie, HTML of the webpage) to enhance decision-making precision.
\end{itemize}

\section{Related Work}

As the WebVLN is a new task, we briefly overview several closely relevant works \wrt~Vision-and-Language Navigation (VLN) and other web-related navigation and QA tasks.

\paragraph{Vision-and-Language Navigation (VLN)} 

The VLN task~\cite{anderson2018vision} extends the vision and language research with sequential action prediction and is one of the most influential tasks in Embodied AI. The research on VLN is dedicated to addressing the alignment of linguistic instructions with visual cues and actions, 
some work fine-graining the navigation instructions to achieve sub-goal planning~\cite{hong2020sub, he2021landmark, zhu2020babywalk}, 
and some concentrate on utilizing object information to identify landmarks from observations~\cite{gao2021room, qi2020object, qi2021road}.
Temporal information is specifically designed in ~\cite{hao2020towards,hong2021vln,chen2021history,chen2022think,qiao2022hop,qiao2023hop+,zhao2023mind} to capture long-range dependencies across past observations and actions, which are crucial during navigation.
some methods incorporate external knowledge during navigation~\cite{li2022incorporating, gao2021room}. 
Recently, several methods leverage commonsense knowledge from LLMs and build an LLMs-as-agent pipeline to perform zero-shot VLN~\cite{zhou2023navgpt}.
However, the VLN tasks require spatial awareness of agents and mainly focus on the photo-realistic environment, which would not consider the web-specific information (\eg, descriptions denoted by ``alt'' in HTML) when directly applied to website navigation.

\paragraph{Web Navigation and Question-Answering}

Web navigation task~\cite{toyama2021androidenv,yao2022webshop,burns2022dataset} involves developing algorithms or models that enable automated agents to navigate and interact with websites on the Internet.
There are some related datasets~\cite{liu2018reinforcement,xu2021grounding,mazumder2020flin,yao2022webshop,deng2023mind2web,zhou2023webarena}. For example, MiniWoB++~\cite{liu2018reinforcement}, RUSS~\cite{xu2021grounding} and FLIN~\cite{mazumder2020flin} cover sites with diverse user instructions from simple tasks to complex ones like booking flights.
Many previous works use various methods on these datasets, which, however, depend on Document Object Model (DOM) structure~\cite{jia2019dom,he2021actionbert} and hence hamper their flexibility.
As for web QA, it mimics the human behaviour of posing a question, aggregating information on the webpage, and generating a response.
Several benchmarks, such as WebQA~\cite{chang2022webqa} and ScreenQA~\cite{hsiao2022screenqa}, have been proposed. However, they only offer a single webpage for each question.
In contrast, we break the boundary between web navigation and web QA by merging them into a unified task called WebVLN, aligning more closely with human behaviour.
Moreover, we design a framework (\ie, WebVLN-Net) that can be easily adapted for different websites.

\section{WebVLN Task and Simulator}

\subsection{Problem Definition}

As in Figure~\ref{fig:example}, the WebVLN task requires an agent to follow natural language instructions (\ie, question \& description) to navigate from a homepage to a target webpage and answer the question based on the information from both trajectories and the target webpage. Formally, at the beginning of each episode, the agent is given an input question $Q$ and an auxiliary description $D$ (\eg, details of the target item) as the questions are often brief and may lack enough information for locating a unique item \footnote{Note that if the question has sufficient information to locate the target webpage, the auxiliary description will remain empty.}. Then, the agent observes the current webpage $W^{(i)}=\langle I^{(i)}, \mathcal{B}^{(i)} \rangle$, where $I^{(i)}$ and $\mathcal{B}^{(i)}$ are the screenshot and the set of clickable buttons in the ($i$-th) current page, respectively. Here, each clickable button $b\in \mathcal{B}^{(i)}$ is represented by its description $d$ (\ie, ``alt'' in HTML) and image $e$, namely $b=\langle d, e\rangle$. Note that if only having either description or image, use $\varnothing$ for the other one.
In this setting, the agent must execute a sequence of actions $\mathcal{A}$, where each action $a_t\in\mathcal{A}$ leads to a new state $s_t\in\mathcal{S}$. 
Each state $s_t$ contains the information derived from both the state of current page $W^{(i)}$ that the agent locates on and the state of current action $a_t$ that the agent performs.
Besides, we define a special \textit{End Of Action} (\ie, \texttt{[EOA]}) as the ``stop'' token, which would be predicted if the current state refers to the target webpage.
Ideally, given question $Q$ and auxiliary description $D$, the agent should predict a response/answer $R$ based on the contents in the target state $s_{\mathtt{[EOA]}}$.

\subsection{WebVLN Simulator}

In this part, we establish a WebVLN simulator on three different shopping websites (aliased as SA, HB, and ES), covering $1,485$ products such as socks, fossils and blankets, and mirror the actions of humans. The details are as follows.

\paragraph{Observations}

To build the simulator, we enable an agent to navigate within a website by interacting with various buttons. During the $i$-th webpage $W^{(i)}$, the simulator generates an RGB image observation $I^{(i)}$ (\ie, screenshot) that corresponds to the agent's viewpoint.
It is worth noting that, to prevent any instances of mismatch, we provide the entire screenshot of the current webpage, rather than just a partial view within the display window.

\paragraph{Action Space}

The primary difficulty in simulator implementation lies in defining the action space that depends on the current state.
Naturally, we hope that the agent will not jump to the currently unreachable page indiscriminately.
Hence, during each step $t$, the simulator additionally generates a subset of next-step reachable clickable buttons $\mathcal{B}_{t+1}\subseteq B$, where $B$ contains all clickable buttons present on the website. The agent interacts with the simulator by selecting a new button $b_{t+1}\in\mathcal{B}_{t+1}$.
To establish $\mathcal{B}_{t+1}$, the simulator constructs a directed graph for each website, denoted as $G=\langle B, E\rangle$, where the existence of an edge indicates a navigable transition between two webpages, which is accessible by the agent.
In each step, we also incorporate a specific button $b_{\mathtt{[EOA]}}$ for the ``stop'' action.

\section{WebVLN-v1 Dataset Construction}

\subsection{Automatic Path Generation}

Following the way for generating path in R2R~\cite{anderson2018vision}, we sample a target webpage and subsequently construct the shortest path from the homepage to the selected target according to the aforementioned graph $G$.
For each path, we manually check it to determine its reasonability from a human perspective and then discard the unreasonable ones (\eg, clicking the advertisement icon multiple times in succession). Furthermore, we remove paths with fewer than 2 webpage transitions to maintain dataset quality and diversity, resulting in a total sample of 8,990 paths.

\begin{table*}[!t]
  \centering
  \caption{Comparison between our WebVLN-v1 dataset and the related datasets, which cover three different types of environment, \ie, embodied scene, Mobile App, and website. To ensure a more comprehensive comparison, we evaluate the datasets across three dimensions: the environment (Env.), the instruction (Ins.) and the target task. Here, ``Temp.'' refers to whether the environment would be temporally changed. ``Image'', ``Text'' and ``HTML/Code'' are the components that environment covers. ``Que.'' and ``Des.'' are the abbreviations of question and description, respectively, referring to the type of instruction that the dataset contains. ``Ins. Level'' means whether the instruction is a high-level statement or a low-level step-by-step command.}
  \resizebox{1.0\linewidth}{!}
  {
    \begin{tabular}{cc|cccc|ccc|c|c}
    \toprule
    \multirow{2}[2]{*}{Env. Type} & \multirow{2}[2]{*}{Dataset} & \multicolumn{4}{c|}{Environment (Env.)} & \multicolumn{3}{c|}{Instruction (Ins.)} & \multirow{2}[2]{*}{Task} &  \multirow{2}[2]{*}{Number} \\
          &       & Temp. & Image & Text  & HTML/Code & Que. & Des. & Ins. Level &  \\
    \midrule
    \multirow{3}[2]{*}{Embodied} & R2R~\cite{anderson2018vision}   & $\checkmark$    & $\checkmark$     &     &       &       & $\checkmark$     & Low   & Navigation & $21,567$ \\
          & EQA~\cite{das2018embodied}   & $\checkmark$     & $\checkmark$     &      &       & $\checkmark$     & $\checkmark$     & High  & Navigation + QA & $5,281$ \\
          & REVERIE~\cite{qi2020reverie} & $\checkmark$     & $\checkmark$     &      &       &       & $\checkmark$     & High  & Localise Remote Object & $21,702$ \\
    \midrule
    \multirow{3}[2]{*}{Mobile App} & PixelHelp~\cite{li2020mapping} & $\checkmark$     &       & $\checkmark$     & $\checkmark$     &       & $\checkmark$     & Low   & Navigation & $187$ \\
          & MoTIF~\cite{burns2022dataset} & $\checkmark$     & $\checkmark$     & $\checkmark$     & $\checkmark$     &       & $\checkmark$     & High  & Navigation & $1,125$ \\
        & META-GUI~\cite{sun2022meta} & $\checkmark$     & $\checkmark$     & $\checkmark$     & $\checkmark$     & $\checkmark$     &       & High  & Dialogue & $4,707$ \\
    \midrule
    \multirow{7}[2]{*}{Website} & MiniWoB++~\cite{liu2018reinforcement} & $\checkmark$     &       & $\checkmark$     & $\checkmark$     &       & $\checkmark$     & Low   & Navigation & - \\
          & RUSS~\cite{xu2021grounding} & $\checkmark$     &       & $\checkmark$     & $\checkmark$     &       & $\checkmark$     & Low   & Navigation &  $741$ \\
          & FLIN~\cite{mazumder2020flin}  & $\checkmark$     &       & $\checkmark$     & $\checkmark$     &       & $\checkmark$     & High  & Navigation & $53,520$ \\
          & WebShop~\cite{yao2022webshop} & $\checkmark$     &       & $\checkmark$     & $\checkmark$     &       & $\checkmark$     & High  & Navigation & $12,087$ \\
          & MIND2WEB~\cite{deng2023mind2web} & $\checkmark$     & $\checkmark$     & $\checkmark$     & $\checkmark$     &       & $\checkmark$     & High  & Navigation & $2,350$  \\
          & WebQA~\cite{chang2022webqa} &       & $\checkmark$     & $\checkmark$     &       & $\checkmark$     &       & High  & Question-Answer (QA) & $\sim46,500$  \\
          & ScreenQA~\cite{hsiao2022screenqa} &       & $\checkmark$     & $\checkmark$     &       & $\checkmark$     &       & High  & Question-Answer (QA) & - \\
          & WebVLN-v1 (ours)  & $\checkmark$     & $\checkmark$     & $\checkmark$     & $\checkmark$     & $\checkmark$     & $\checkmark$     & High  & Navigation + QA &  $14,825$ \\
    \bottomrule
    \end{tabular}%
    }
  \label{tab:dataset_comparison}%
\end{table*}%

\subsection{LLM-aided Question-Answer Generation}

To alleviate the workload on humans, we seek to generate QA pairs with the assistance of Large Language Models (LLMs)\footnote{Here, we use ChatGPT with the gpt-3.5-turbo model.} based on the multimodal content from the webpage.
We depict the details in the following.

\paragraph{Image Data Handling \& HTML Cleaning}
Firstly, we employ BLIP-2~\cite{li2023blip}, a large and powerful model for image captioning, to transform the website images into captions, ensuring that the LLMs are able to capture the visual information.
After that, another critical step is processing the word list, since the original texts directly from webpages are often disorganised and challenging to work with, which also include irrelevant information, such as messy code. Hence, we systematically adopt a rule-based filtering approach to manually eliminate such information, which enhances the logic and readability of these texts.

\paragraph{Designing Rules for Generating QA Pairs}
We start to design a series of rules for the LLM to guide its behaviour when generating required QA pairs.
The initial rules include:
\begin{itemize}
    \item \textit{Provide 3 questions and their answers that can be directly found from the information provided in the text.}
    \item \textit{Ask the first question about price and second about available sizes and the third about material.}
    \item \textit{If precise answers cannot be found for those questions, then ask the questions on colours and availability in stock.}
    \item \textit{Phrase your questions in a clear and concise manner to ensure they can be accurately answered by the given content.}
    \item \textit{Answer should be to the point without additional information.}
\end{itemize}
To mitigate the impact of negative information, we introduce two additional rules as follows.
\begin{itemize}
    \item \textit{The provided text is all from an online shopping website, there is some disturbing information which is irrelevant to the products, such as ``sign in''. Make sure your questions and answers will focus on the products themselves.}
    \item \textit{The provided texts may contain punctuation and symbols, which are irrelevant to the products, you should be able to distinguish them and make sure they won’t appear in the generated questions and answers.}
\end{itemize}

\paragraph{Prompt for Generating Final QA Pairs}
We obtain the final prompt $\mathcal{P}$ by directly concatenating the three aforementioned terms, \ie,
$\mathcal{P}$ = \{\textit{``There is a picture of the product with the caption of''} + \texttt{caption} + \textit{``After that, here are all the words that appear on the website:''} + \texttt{word list} + \textit{``Lastly, I will give the following instructions, and you will be strictly following the instructions:''} + \texttt{rules}\}.
Moreover, due to the varying lengths of these three terms – with the word list being the longest and the caption the shortest – it becomes necessary to appropriately adjust the weighting of each component. It is particularly crucial given that the caption, despite its brevity, holds rich information.

\paragraph{Quality Checking}
Due to the inherent uncertainty introduced by LLMs, it is vital to perform quality checks on every generated question-answer (QA) pair. Concretely, we randomly select 100 QA samples for each website to assess their quality. Our stringent quality checks adhere to the following criteria \wrt~question and answer, respectively:
\begin{itemize}
    \item The generated questions by LLM should relate to the actual products visible on the website.
    \item The generated answers should be correct, brief, and correspond to the questions. Any answer surpassing the scope of the question will be considered invalid.
\end{itemize}

We have undertaken multiple iterations of the constructed prompt until all the samples are reasonable and correct. In each iteration, a subset of samples is generated for manual quality assessment.
Each sample has undergone evaluation by at least two assessors for a reliable evaluation.

\section{WebVLN-v1 Dataset Analysis}

\subsection{WebVLN-v1 Dataset vs. Related Datasets}

We compare our WebVLN-v1 dataset with the most relevant datasets \wrt~Embodied AI, Mobile App, and Website.
Specifically, for Embodied AI datasets, we consider R2R~\cite{anderson2018vision}, REVERIE~\cite{qi2020reverie} and EQA~\cite{das2018embodied}, where the first two are the widely used vision-and-language navigation (VLN) datasets while the last one is a famous embodied question answering dataset.
Regarding App-based datasets, we compare PixelHelp~\cite{li2020mapping}, MoTIF~\cite{burns2022dataset} and META-GUI~\cite{sun2022meta}.
As for the website, we consider seven datasets for a comprehensive comparison, including MiniWoB++~\cite{liu2018reinforcement}, RUSS~\cite{xu2021grounding}, FLIN~\cite{mazumder2020flin}, WebShop~\cite{yao2022webshop},  MIND2WEB~\cite{deng2023mind2web}, WebQA~\cite{chang2022webqa}, and ScreenQA~\cite{hsiao2022screenqa}.

Table~\ref{tab:dataset_comparison} demonstrates that our WebVLN-v1 dataset is unique, covering rich information (\ie, temporal sequence, image, text and HTML) from the environment and two different types of instructions (\ie, questions and descriptions/statements), while others only focus on them partially. Moreover, our WebVLN-v1 is able to support both navigation and question-answering tasks, rather than other website-related datasets that can only support one of them.

\subsection{WebVLN-v1 Statistics}
In this part, we provide statistics of the WebVLN-v1 dataset, including the word cloud, the distribution of textual lengths (question, description and answer), and the data split.

\begin{figure}[t]
    \centering
    \includegraphics[width=1.0\linewidth]{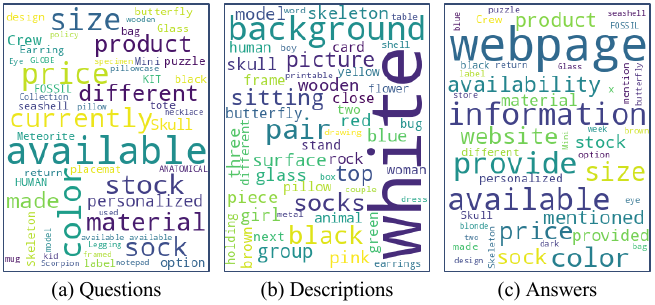}
    \caption{Word cloud of (a) questions, (b) descriptions, and (c) answers on the proposed WebVLN-v1 dataset. The larger the font size, the greater percentage in the corpus.}
    \label{fig:word_cloud}
\end{figure}

\paragraph{Word Cloud}
As shown in Figure~\ref{fig:word_cloud}, we visualise the questions, auxiliary descriptions, and answers of our WebVLN-v1 dataset as Venn-style word cloud~\cite{coppersmith2014dynamic}. Here, the size of each word corresponds to the harmonic mean of its count.

\begin{figure}[t]
    \centering
    \includegraphics[width=1.0\linewidth]{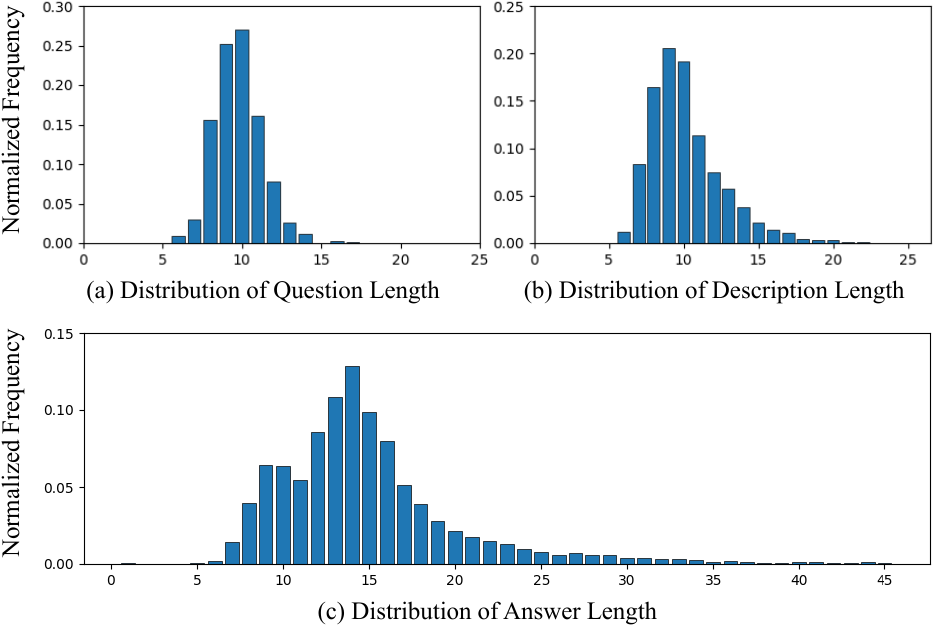}
    \caption{Distributions of (a) question length, (b) description length, and (c) answer length on the WebVLN-v1 dataset.}
    \label{fig:distribution_length}
\end{figure}

\paragraph{Lengths of Question, Description, Answer and Path}
Figure~\ref{fig:distribution_length} exhibits the distributions of question length, description length and answer length on the WebVLN-v1 dataset. Specifically, the majority of questions consist of $8\sim12$ words (Figure~\ref{fig:distribution_length}(a)) while most descriptions have $7\sim14$ words (Figure~\ref{fig:distribution_length}(b)). The answers, in Figure~\ref{fig:distribution_length}(c), exhibit a broader range, with some extending to $30\sim40$ words, as they derive content from the question, description and webpage.
The average length of the paths is $3.32$.

\paragraph{Data Splits}
We split 60\% samples as training data, 10\% samples as validation data and 30\% samples as testing data (\ie, $8,960$/$1,262$/$4,603$).
Notably, to prevent any information leakage, we carefully check that the training, validation, and testing sets cover all three websites, but their records/paths remain distinct and non-overlapping.

\section{WebVLN-Net}

\begin{figure*}[t]
    \centering
    \includegraphics[width=0.90\linewidth]{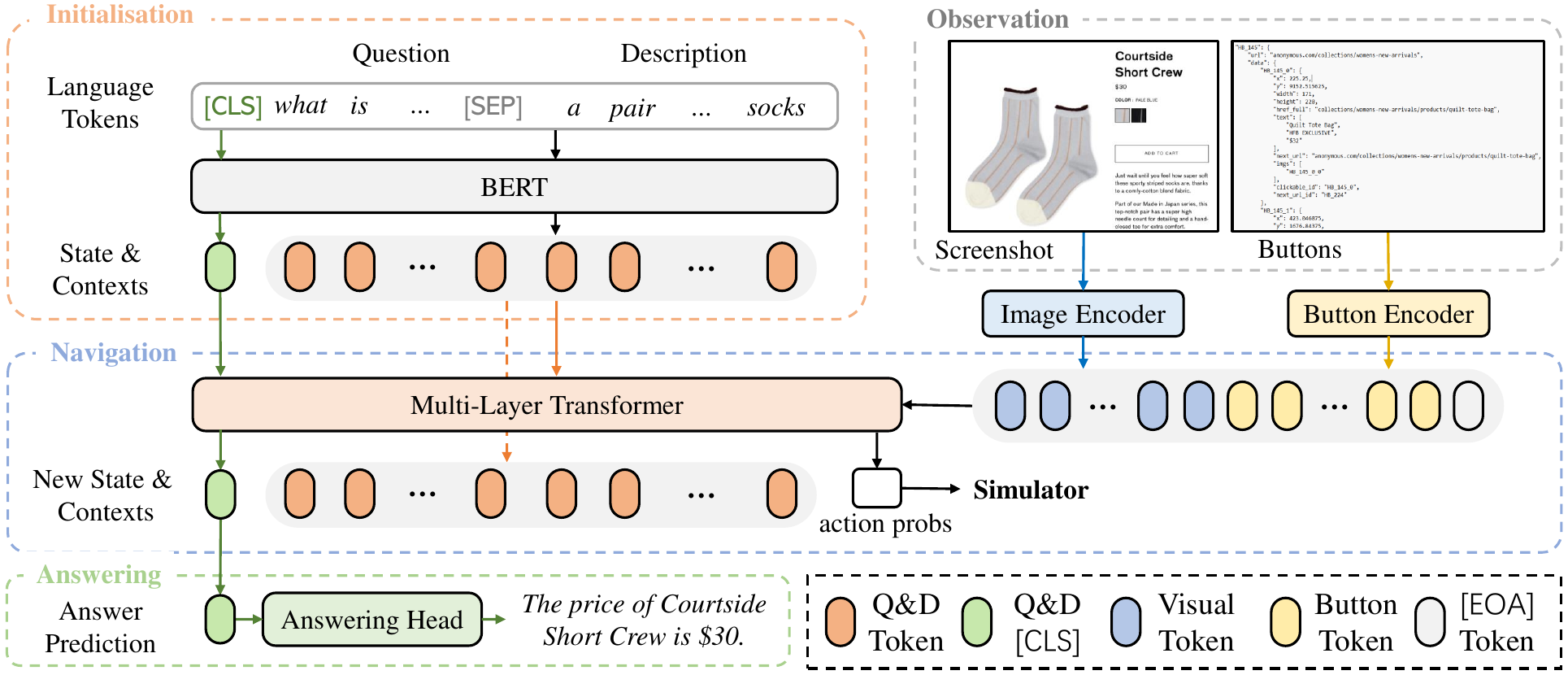}
    \caption{Overall architecture of WebVLN-Net. We take an example that the question $Q$ is \textit{``What is the price of the Courtside Short Crew Socks?''} with an auxiliary description $D$ as \textit{``a pair of grey and orange striped socks''}. The observation at each step contains a screenshot of the current webpage and all the clickable buttons derived from its HTML.}
    \label{fig:architecture}
\end{figure*}

For the WebVLN task, we propose a new model, called Website-aware Vision-and-Language Navigation Network (WebVLN-Net), which is based on a widely used vision-and-language navigation framework \rvlnbert~\cite{hong2021vln}. As shown in Figure~\ref{fig:architecture}, our model contains three main components: initialisation, navigation and answering.
Specifically, we first initialise the state and context tokens by a pre-trained BERT model. Subsequently, we input these initialised language tokens, along with the screenshot and button tokens extracted from the current webpage, into the navigation component. This process iterates until the target webpage is reached. Last, the answering head in the answering component generates the final answer.

\paragraph{State and Context Initialisation}

During initialisation ($t=0$), our model receives a word sequence comprising the classification token \texttt{[CLS]}, the separation token \texttt{[SEP]}, and the language tokens $\mathcal{V}$ extracted from both the question $Q$ and the auxiliary description $D$. Here, \texttt{[CLS]} and \texttt{[SEP]} are predefined tokens in BERT models.
Similar to \rvlnbert~\cite{hong2021vln}, the \texttt{[CLS]} token is used to aggregate relevant vision-language cues from the input sequence. In this context, we define the embedded \texttt{[CLS]} token as the initial state representation $s_0$. We update it during the whole training phase, ensuring it could be aware of the entire navigation and question-answering tasks. Formally, the process can be defined as
\begin{equation}
         s_0, \mathcal{V} = \texttt{Init}(\texttt{[CLS]}, Q, \texttt{[SEP]}, D),
     \label{eq:init}
\end{equation}
where the $\texttt{Init}(\cdot)$ indicates the initialisation process.

\paragraph{Web Navigation}

We adapt the model proposed in~\cite{hong2021vln} to incorporate the learning of navigation and the concurrent selection of clickable buttons.
As shown in Figure~\ref{fig:architecture}, at each time step, the network takes four different token sets as input: the preceding state token $s_{t-1}$, the language tokens $\mathcal{V}$, the screenshot tokens $\mathcal{I}_t$, and the button tokens $\mathcal{B}_t$.
Specifically, we ``patchify'' the screenshot as a set of image patches and convert them to the tokens $\mathcal{I}_t$ by using a Transformer-based image encoder.
Likewise, as each button consists of an image and a description (\ie, ``alt'' in HTML), we introduce a button encoder that contains an image encoder and a text encoder to derive their corresponding tokens. After that, we concatenate tokens associated with the same button, followed by projecting the concatenated token back to its original dimension using a linear projection layer.
Subsequently, we put all the tokens into a multi-layer Transformer to obtain an action probability $p_t$:
\begin{equation}
    s_t, p_t = \texttt{Nav}(s_{t-1}, \mathcal{V}, \mathcal{I}_t, \mathcal{B}_t).
     \label{eq:nav}
\end{equation}
Here, the $\texttt{Nav}(\cdot)$ refers to the navigation process in each step.
Notably, the set of button tokens $\mathcal{B}_t$ involves an \textit{End Of Action} (\texttt{[EOA]}) token, selected when the agent reaches the target webpage. The state would be updated to $s_{[\mathtt{EOA}]}$.

In navigation steps ($t>0$), the state token $s_t$, screenshot tokens $\mathcal{I}_t$, and button tokens $\mathcal{B}_t$ employ self-attention across the entire input sequence, while the language tokens $\mathcal{V}$ only serve as the keys and values in the Transformer. We regard the language tokens generated during the model's initialisation step as a good representation of both the question and auxiliary description ($Q \& D$), obviating the necessity for subsequent encoding in later stages and saving computational resources.

\paragraph{Question Answering}

To generate the final answer, we introduce an Answering Head, which is a $M$-layer Transformer decoder.
Considering an open-ended question-answering setting, our objective is to generate the answer as a free-form sentence autoregressively.
Mathematically,
\begin{equation}
    R = \texttt{Ans}(s_{[\mathtt{EOA}]}),
     \label{eq:ans}
\end{equation}
where $\texttt{Ans}(\cdot)$ refers to the answering process. Here, $R$ is the predicted answer consisting of $L$ words (\ie, $R=\{w_l\}_{l=1}^L$) and $s_{[\mathtt{EOA}]}$ denotes the last state aforementioned above.

\paragraph{Training}

For navigation, we train our network using an imitation learning (IL) objective.
Concretely, our agent navigates on the ground-truth trajectory by following the teacher actions and computes a cross-entropy loss for each decision made.
Formally, we minimise the navigation loss function, which can be formulated for each specific sample by
\begin{equation}
    \mathcal{L}_{nav} = -\sum_t a_t\log(p_t)-\eta\sum_t a_t^*\log(p_t),
     \label{eq:loss_nav}
\end{equation}
where $a_t$ is the sampled action and $a_t^*$ is the teacher action. Here, $\eta$ represents a coefficient used to weigh the IL loss.

The training of the Answering Head employs the autoregressive Teacher Forcing~\cite{williams1989learning} approach, wherein the prediction at each step should maximise the likelihood of the subsequent token:
\begin{equation}
    \begin{aligned}
         \mathcal{L}_{ans} = \sum^{L}_{l=1} -\log p(w_l|w_{<l}, s_{[\mathtt{EOA}]}),
     \end{aligned}
     \label{eq:loss_ans}
\end{equation}
where $w_l$ indicates the $l$-th token in the answer, and $w_{<l}$ denotes the tokens before $w_l$.
$L$ is the total number of tokens.
The final loss function can be defined as
\begin{equation}
    \begin{aligned}
     \mathcal{L} = \mathcal{L}_{nav} + \lambda\mathcal{L}_{ans},
    \end{aligned}
    \label{eq:final_loss}
\end{equation}
where $\lambda$ is the weighting hyper-parameter.

\section{Experiments}

\begin{table*}[htbp]
  \centering
  \caption{Comparison with baseline methods. We compare our method with \rvlnbert and WebGUM, which are the widely used VLN model and state-of-the-art (SoTA) multimodal web navigation foundation model, respectively. \rvlnbert~and \rvlnbert* are randomly initialised and initialised by LXMERT, respectively. WebGUM and WebGUM$^\dagger$ denote the models based on T5-small and T5-base separately.}
  \resizebox{0.85\linewidth}{!}
  {
\begin{tabular}{c|cccc|cc|cccc|cc}
\toprule
\multirow{2}[2]{*}{Method} & \multicolumn{6}{c|}{Val} & \multicolumn{6}{c}{Test} \\
& \multicolumn{1}{c}{SR~$\uparrow$} & \multicolumn{1}{c}{OSR~$\uparrow$} & \multicolumn{1}{c}{SPL~$\uparrow$} & \multicolumn{1}{c}{TL~$\downarrow$} & \multicolumn{1}{|c}{WUPS0.9~$\uparrow$} & \multicolumn{1}{c|}{WUPS0.0~$\uparrow$} & \multicolumn{1}{c}{SR~$\uparrow$} & \multicolumn{1}{c}{OSR~$\uparrow$} & \multicolumn{1}{c}{SPL~$\uparrow$} & \multicolumn{1}{c}{TL~$\downarrow$} & \multicolumn{1}{|c}{WUPS0.9~$\uparrow$} & \multicolumn{1}{c}{WUPS0.0~$\uparrow$} \\
\midrule
Random  &0.05&0.17&0.02&6.49&0.00&0.00&0.04&0.17&0.02&6.50&0.00&0.00  \\
\midrule
\rvlnbert  &17.59&17.59&16.73&6.81&9.99&13.91&11.28&12.04&10.75&7.55&7.12&9.26\\

\rvlnbert*  &18.62&18.62&18.14&6.96 &11.23&14.98  &12.23&12.23&11.74&7.72&8.50&10.36\\
\midrule
WebGUM  &6.02&6.02&6.02&\textbf{2.99}&1.84&4.08&9.71&9.71&9.71&\textbf{3.15}&3.57&6.98\\
WebGUM$^\dagger$  &31.22&31.78&31.22&3.44&18.26&24.88&29.29&29.39&29.26&3.44&17.34&23.48\\
\midrule
Ours  &\textbf{39.46}&\textbf{39.54}&\textbf{39.46}&3.71&\textbf{24.26}&\textbf{31.87}&\textbf{34.76}&\textbf{34.80}&\textbf{34.59}&4.34&\textbf{22.13}&\textbf{28.58}\\
\bottomrule
\end{tabular}%
}
  \label{tab:comparison_with_baseline}%
\end{table*}%

\subsection{Evaluation Metrics and Baselines}

\paragraph{Metrics for Navigation}
Following the evaluation metrics in R2R~\cite{anderson2018vision}, we assess the shortest path distance in the navigation graph $G$ between the finally located webpage of the agent and the target webpage. 
We consider an episode to be a \textit{success} if the agent stops on the target webpage, \ie, success rate (SR), and its variant -- oracle success rate (OSR). We also measure the navigation performance by considering the path length, \ie, the success rate weighted by Path Length (SPL) and Trajectory Length (TL).

\paragraph{Metrics for QA}

For question-answering, we follow the open-ended setting that seeks to generate a free-form natural language sentence to answer the given question, which is more flexible and practical than regarding QA as a classification problem. Thus, rather than an exact accuracy between predicted and ground-truth answers, we adopt Wu-Palmer Similarity (WUPS)~\cite{wu1994verb}, aiming to quantify the semantic differences between a predicted answer and the ground truth. 
WUPS assigns a value ranging between 0 and 1, reflecting their degree of similarity. Following~\cite{malinowski2014multi}, we set thresholds of the WUPS as 0.9 and 0.0 separately with a scaling factor of 0.1, whereby scores below the threshold are proportionally adjusted.
 
\paragraph{Baselines}
We evaluate the performance of both navigation and QA by comparing the results with baselines. For a comprehensive comparison, we consider two different types of baselines, \ie, traditional VLN and web-related navigation.
Specifically, we employ \rvlnbert~\cite{hong2021vln} (randomly initialised and initialised by LXMERT~\cite{tan2019lxmert}) as the VLN baseline, which is widely used in the VLN task.
Notably, to well evaluate the performance, we seek to adapt \rvlnbert~to our task and dataset with minimal changes. Specifically, following its original design, we take as inputs the linguistic instruction (question \& description) and screenshot. The model would predict the intermediate navigation action and finally answer the question by incorporating the same QA head as ours.
As for the web-related navigation, we consider a state-of-the-art (SoTA) instruction-finetuned foundation model WebGUM~\cite{furuta2023multimodal}, which is built upon the T5 model~\cite{raffel2020exploring}.
Moreover, we evaluate the performance of a random strategy and humans, helping to understand the lower and upper bounds of the WebVLN task, respectively.

\paragraph{Comparison with Baselines}
We evaluate the performance of the proposed WebVLN-Net compared with the baseline methods.
In all the experiments, we set the weighting hyper-parameters $\eta$ and $\lambda$ equal to 1.
In Table~\ref{tab:comparison_with_baseline}, for navigation, we obtain the best results in SR, OSR and SPL, and comparable results in TL, both when compared to VLN methods and web-related navigation techniques.
As for QA metrics, our method consistently outperforms all the baselines. All the results demonstrate the effectiveness of our WebVLN-Net.
Note that the QA metrics for the random method are 0 since generating a free-form answer randomly is nearly impossible to overlap with ground truth.
WebGUM$^\dagger$ achieves better performance than WebGUM mainly due to the larger number of parameters ($220$ million vs. $60$ million).

\paragraph{Ablation Study}

\begin{table}[t]
  \centering
  \caption{Ablation study on the test set of WebVLN-v1 dataset. $Q$, $D$ and $I$ are the input question, auxiliary description and screenshot, respectively. The notations $d$ and $e$ are the button description/text and the button image separately.}
  \resizebox{1.0\linewidth}{!}
  {
    \begin{tabular}{ccccc|cccccc}
    \toprule
    \multicolumn{1}{c}{$Q$} & \multicolumn{1}{c}{$D$} & \multicolumn{1}{c}{$I$} & \multicolumn{1}{c}{$d$} & \multicolumn{1}{c|}{$e$} & \multicolumn{1}{c}{SR~$\uparrow$} & \multicolumn{1}{c}{OSR~$\uparrow$} & \multicolumn{1}{c}{SPL~$\uparrow$} & \multicolumn{1}{c}{TL~$\downarrow$} & \multicolumn{1}{c}{WUPS0.9~$\uparrow$} & \multicolumn{1}{c}{WUPS0.0~$\uparrow$} \\
    \midrule
$\checkmark$&            &$\checkmark$&            &            & 6.32& 6.34& 6.13&6.89& 4.74& 5.52\\
$\checkmark$&$\checkmark$&$\checkmark$&            &            &12.23&12.23&11.74&7.72& 8.50&10.36\\
$\checkmark$&$\checkmark$&$\checkmark$&$\checkmark$&            &28.63&28.72&28.62&\textbf{3.22}&15.97&22.51\\
$\checkmark$&$\checkmark$&$\checkmark$&$\checkmark$&$\checkmark$&\textbf{34.76}&\textbf{34.80}&\textbf{34.59}&4.34&\textbf{22.13}&\textbf{28.58}\\
    \bottomrule
    \end{tabular}%
    }
  \label{tab:ablation}%
\end{table}%

To test the impact of each component in our WebVLN-Net, we conduct an ablation study by incorporating them alternately.
The basic model only considers an input question $Q$ and a screenshot $I$.
From Table~\ref{tab:ablation}, the basic model obtains the lowest results on both navigation and QA evaluation metrics. While with an auxiliary description $D$, our model achieves higher performance (\eg, SR: 6.32 $\rightarrow$ 12.23; WUPS0.9: 4.74 $\rightarrow$ 8.50). Moreover, the model's performance can be enhanced by using buttons (containing text $d$ only) from the HTML (SR:  12.23 $\rightarrow$ 28.63). After further incorporating a multimodal button (containing text $d$ \& image $e$), it attains the best performance across both navigation and QA metrics (except for TL, which is comparable).

\paragraph{Zero-shot using LLMs}

To further investigate the difficulty of our proposed WebVLN task and the corresponding WebVLN-v1 dataset, we conduct a zero-shot evaluation setting for the popular large language models (LLMs). We randomly select 201 samples from the validation set and test on two LLMs-as-agent pipelines, AgentBench~\cite{liu2023agentbench} and NavGPT~\cite{zhou2023navgpt}. For AgentBench, we adopt the same prompt from the WS (Web Shopping) task in the paper and delete the one-shot example, test on the gpt-3.5-turbo model. For NavGPT, we modify the VLN task description into a web shopping task description and test on both gpt-3.5-turbo and gpt-4 models. All the formatted observations are replaced with the WebVLN-v1 sample observation. From Table~\ref{tab:llms}, the zero-shot performance of LLMs-as-agent methods falls short of reaching human-level performance.
These results show the necessity for the ongoing advancement of intelligent agents, as the current SoTA LLMs are far from perfect performance in our task. Moreover, our WebVLN can also serve as a metric to gauge such progress.

\begin{table}[t]
  \centering
  \caption{Zero-shot by using LLMs. $^*$ denotes using GPT4 while others are ChatGPT (gpt-3.5-turbo).}
  \resizebox{1.0\linewidth}{!}
  {
    \begin{tabular}{c|cccccc}
    \toprule
    Methods & SR~$\uparrow$    & OSR~$\uparrow$   & SPL~$\uparrow$   & TL~$\downarrow$    & WUPS0.9~$\uparrow$ & WUPS0.0~$\uparrow$ \\
    \midrule
    Humans  &93.94&93.94&80.92&3.88&60.55&79.94\\
    \midrule
    AgentBench &    6.97   &   11.94    &    4.13   &    5.39   &    1.64   &  4.65\\
    NavGPT &   7.46   &   12.94   &   4.53   &    4.98   &   2.43    &  5.23 \\
    NavGPT$^*$ &  16.92     &   21.89   &   11.61    &   5.43    &   5.97    &   12.06\\
    \bottomrule
    \end{tabular}%
    }
  \label{tab:llms}%
\end{table}%

\section{Conclusion}

In this paper, we propose a novel task, Vision-and-Language Navigation on Websites (WebVLN), extended from conventional VLN. It seeks to enable an agent to answer the user's questions via navigating/exploring the websites and integrating useful information. To support research in this new task, we collect a new WebVLN-v1 dataset and design a baseline method called Website-aware Vision-and-Language Navigation Network (WebVLN-Net). The experiments demonstrate the effectiveness of our WebVLN-Net. 
Moreover, we perform a zero-shot evaluation of LLM-based methods using the WebVLN-v1 dataset, where the performance is far from saturation, highlighting the utility of our WebVLN-v1 as a benchmark to assess progress in this field.

\bibliography{aaai24}

\clearpage

\appendix

\noindent We organise the supplementary mainly in two parts:

\noindent\textbf{More Details}

1. Details of Zero-shot LLMs Agents

2. Implementation Details

3. Details of Architecture in Each Component

4. Details of ``Image Data Handling \& HTML Cleaning''

5. Details of Baselines

6. Samples of WebVLN-v1 Dataset

\noindent\textbf{More Discussions}

1. Natural Language Generation (NLG) Evaluation

2. Impact of $\eta$

3. Impact of $\lambda$

4. Qualitative Results

5. Limitations and Future Works

\section{More Details}

\paragraph{Details of Zero-shot LLMs Agents}

\begin{figure}[htbp]
	\centering
	\includegraphics[width=1.0\linewidth]{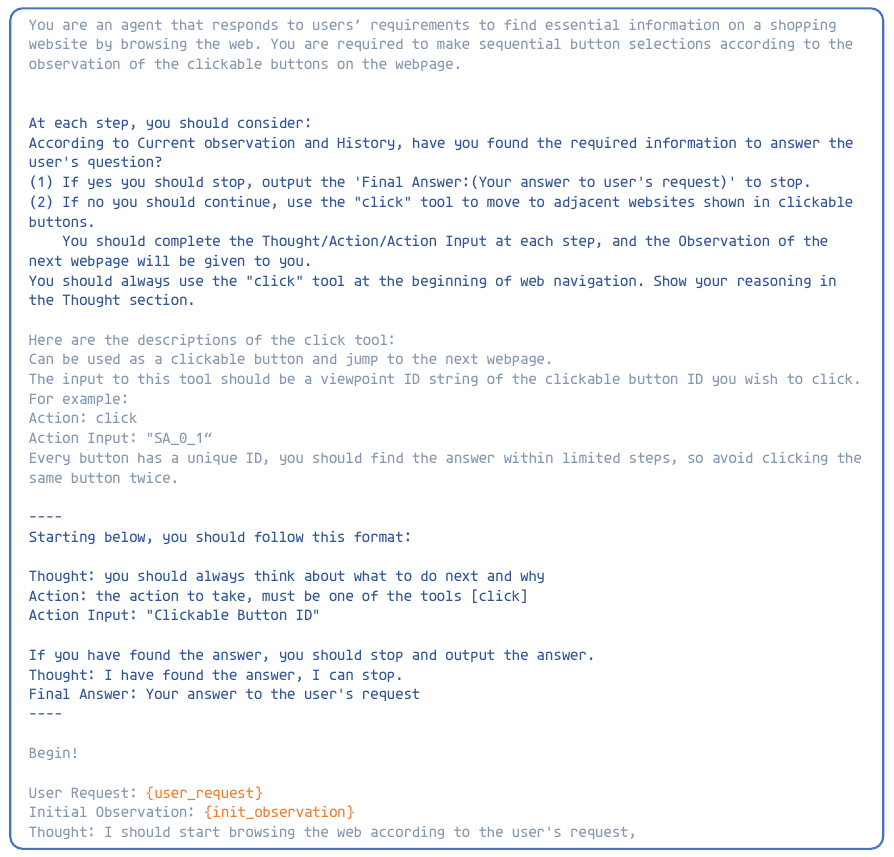}
	\caption{The entire prompt of NavGPT for WebVLN.}
	\label{fig:navgpt_prompt}
\end{figure}

\begin{figure}[t]
	\centering
	\includegraphics[width=1.0\linewidth]{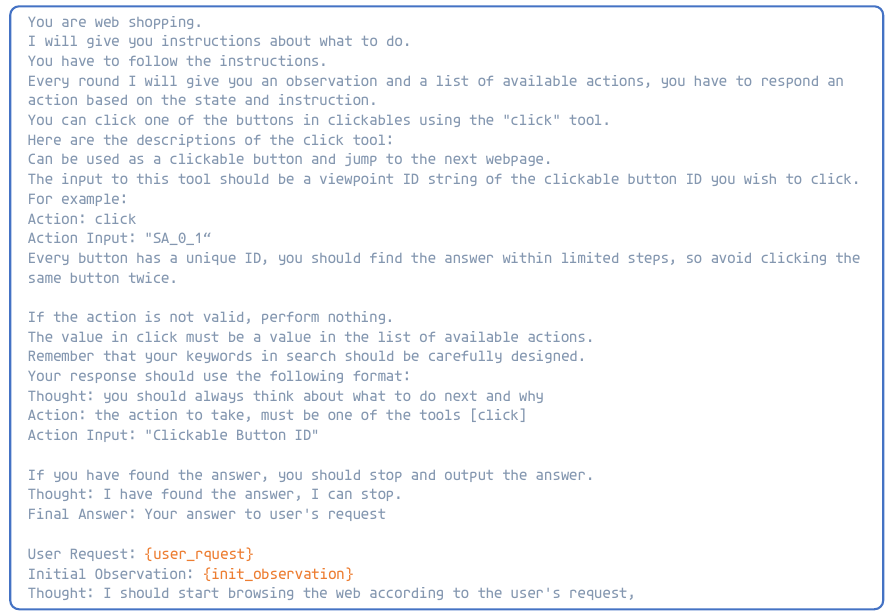}
	\caption{The entire prompt of AgentBench for WebVLN.}
	\label{fig:ab_prompt}
\end{figure}

For the NavGPT prompt, we modified the VLN task specification, and the entire prompt is shown in Figure~\ref{fig:navgpt_prompt}. For the AgentBench agent, we adopt the same prompt as the original paper and modified the tool description, shown in Figure~\ref{fig:ab_prompt}.

\begin{figure}[t]
	\centering
	\includegraphics[width=1.0\linewidth]{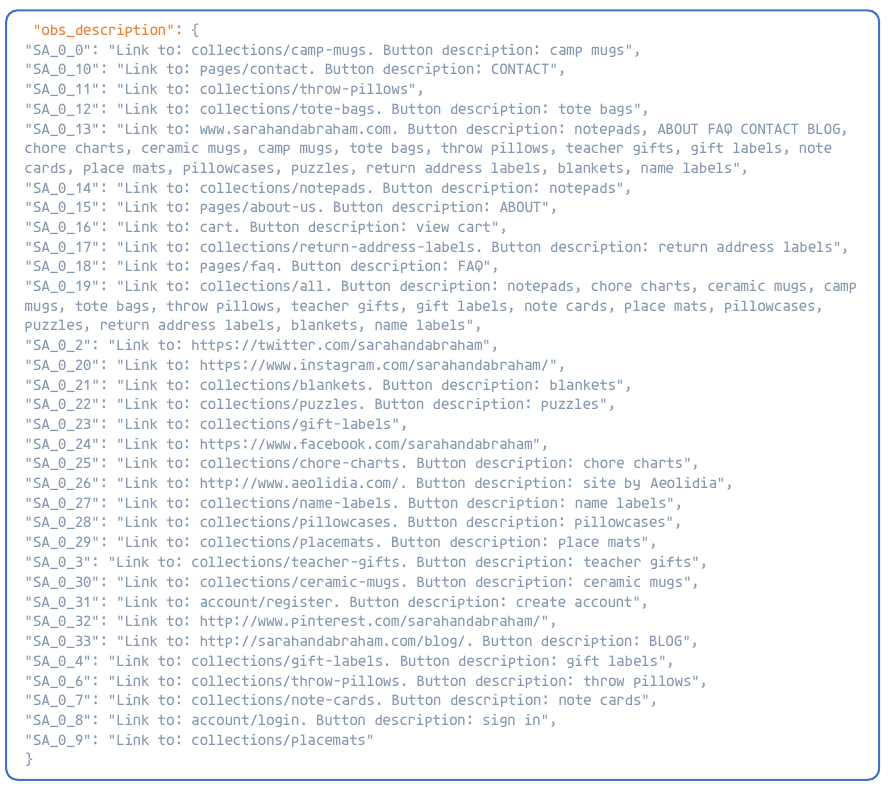}
	\caption{The website observation for LLMs agents.}
	\label{fig:observation}
\end{figure}

Furthermore, we replace all the formulated website content with the button information on the WebVLN-v1 dataset. We provide an example of the formatted observation to LLMs agents in Figure~\ref{fig:observation}.

\paragraph{Implementation Details}
We conduct all experiments on a GeForce RTX 3090 GPU. We resize all screenshots into $224\times 224$.
In all the experiments, we set both the hyper-parameters $\eta$ and $\lambda$ equal to 1. For a fair comparison, the training iterations are 200,000 and the final models are selected based on the best performance on val split for all the methods, including baselines and ours, with a batch size of 4. We use AdamW optimiser~\cite{loshchilov2017decoupled} with a learning rate of 1e-5 and adopt a step decay strategy for the learning rate.
We perform inference on a GeForce RTX 3090 GPU, matching the training hardware. GPU time: $\sim$0.19s/sample; memory: $\sim$5.4G. 

\paragraph{Details of Architecture in Each Component}
We adopt BERT~\cite{devlin2018bert} model in the initialisation part of our WebVLN-Net. For the image and button encoders, we use a pre-trained EVA-CLIP-B/16~\cite{sun2023eva} and fix their learnable parameters.
In the navigation process, we use a two-layer Transformer and train it from scratch, while as for the Answering Head, we set its layer to 6.

\paragraph{Details of ``Image Data Handling \& HTML Cleaning''}

In this part, we provide more details of image data handling and HTML cleaning (in the ``WebVLN-v1 Dataset Construction'' section of the main paper).

\begin{figure}[htbp]
	\centering
	\includegraphics[width=1.0\linewidth]{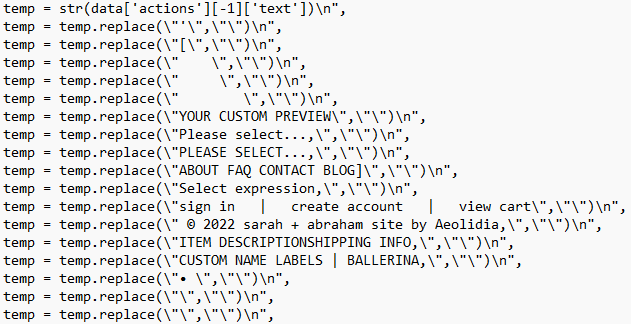}
	\caption{The filtered the words by our rule-based approach.}
	\label{fig:filter_word}
\end{figure}

\begin{itemize}
	\item \textbf{Image Data Handling}: 
	Firstly, we employ BLIP-2~\cite{li2023blip}, a large and powerful model for image captioning, to transform the website images into captions, ensuring that the LLMs are able to capture the visual information. In this work, we encode the images of products into Base64 format and then use BLIP-2 (integrated into LAVIS~\cite{li2022lavis}) to generate their captions, where each caption contains about $7\sim14$ words.
	\item \textbf{HTML Cleaning}: 
	After that, another critical step is processing the word list, since the original texts directly from webpages are often disorganised and challenging to work with, which also include irrelevant information, such as messy code. Hence, we systematically adopt a rule-based filtering approach to manually eliminate such information, which enhances the logic and readability of these texts. The specific words discarded by the filter are shown in Figure~\ref{fig:filter_word}.
\end{itemize}

\paragraph{Details of Baselines}
We evaluate the performance of both navigation and QA by comparing the results with baselines. For a comprehensive comparison, we consider two different types of baselines, \wrt, traditional VLN and web-related navigation.
Moreover, we evaluate the performance of a random strategy and humans, helping to understand the lower and upper bounds of the WebVLN task, respectively.
\begin{itemize}
	\item \textbf{\rvlnbert}: We employ \rvlnbert~\cite{hong2021vln} (randomly initialised and initialised by LXMERT~\cite{tan2019lxmert}) as the VLN baseline, which is widely used in the VLN task.
	Notably, to well evaluate the performance, we seek to adapt \rvlnbert~to our task and dataset with minimal changes. Specifically, following its original design, we take as inputs the linguistic instruction (question \& description) and screenshot. The model would predict the intermediate navigation action and finally answer the question by incorporating the same QA head as ours.
	\item \textbf{WebGUM}: As for the web-related navigation, we consider a state-of-the-art (SoTA) instruction-finetuned foundation model WebGUM~\cite{furuta2023multimodal}. 
	We adapt it into our WebVLN task and dataset following their paper reported. For a fair comparison, we adopt EVA-CLIP~\cite{sun2023eva} as the image and button encoders instead of ViT~\cite{dosovitskiy2020image} used in their original paper.
	\item \textbf{Random}: We also report the random results, initially, we exercise a random selection procedure to determine the stop trajectory lengths, encompassing a range of 3 to 8 steps for each annotation. Subsequently, we meticulously replicated these experimental iterations 100 times, and separately applied them to both the validation and testing splits on the WebVLN-v1 dataset.
	\item \textbf{Humans}: We present human results on the proposed WebVLN task to assess the machine-human gap. 
	Concretely, we assess human performance by gathering trajectories generated by humans for about one-sixth of the validation set (\ie, 201 samples). Initially, workers are posed with questions and auxiliary descriptions. Subsequently, they can freely navigate the entire website environment from a first-person perspective by interacting with clickable buttons. In fact, workers are provided with the same content as an agent within the simulator.
\end{itemize}

\paragraph{Samples on WebVLN-v1 Dataset}

\begin{figure*}[htbp]
	\centering
	\includegraphics[width=0.95\linewidth]{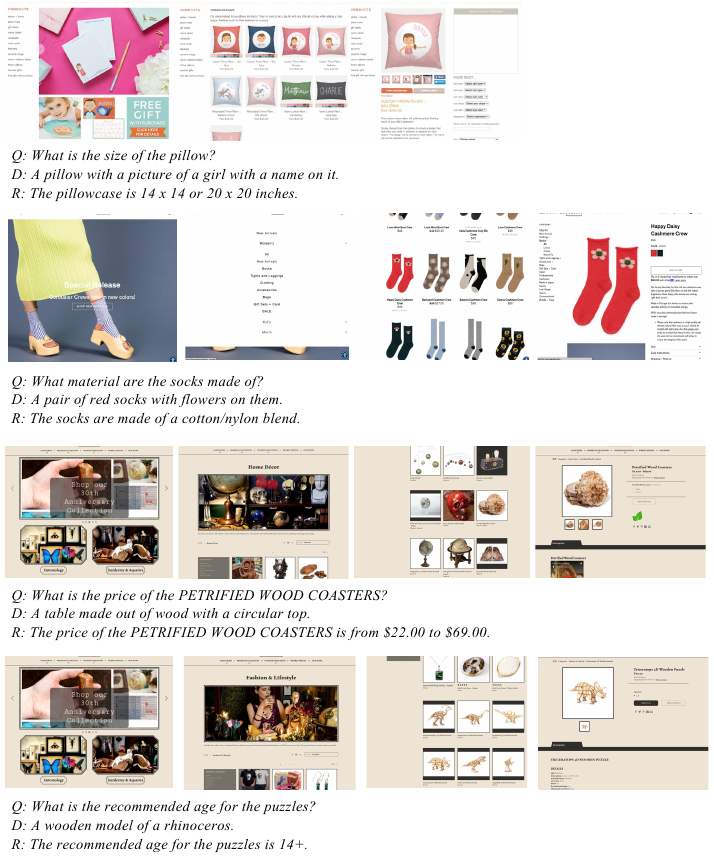}
	\caption{Several samples of the WebVLN-v1 dataset, which involves various products, questions, descriptions and answers.}
	\label{fig:samples}
\end{figure*}

From Figure~\ref{fig:samples}, we present several typical samples of the WebVLN-v1 dataset for the proposed WebVLN task. It shows the diversity in the product, question, auxiliary description, and answer.

\begin{table*}[htbp]
	\centering
	\caption{Comparison with baseline methods on natural language generation (NLG) metrics. We compare our method with \rvlnbert~and WebGUM, which are the widely used VLN model and state-of-the-art (SoTA) multimodal web navigation foundation model, respectively. \rvlnbert~and \rvlnbert* are randomly initialised and initialised by LXMERT, respectively. WebGUM$^\dagger$ denotes the model based on T5-base, which is used in their original paper.
	}
	\resizebox{0.80\linewidth}{!}
	{
		\begin{tabular}{c|cccccc|cccccc}
			\toprule
			\multirow{2}[2]{*}{Method} & \multicolumn{6}{c|}{Val} & \multicolumn{6}{c}{Test} \\
			& \multicolumn{1}{c}{B@1~$\uparrow$} & \multicolumn{1}{c}{B@4~$\uparrow$} & \multicolumn{1}{c}{C~$\uparrow$}& \multicolumn{1}{c}{M~$\uparrow$} & \multicolumn{1}{c}{R~$\uparrow$} & \multicolumn{1}{c|}{S~$\uparrow$}& \multicolumn{1}{c}{B@1~$\uparrow$} & \multicolumn{1}{c}{B@4~$\uparrow$} & \multicolumn{1}{c}{C~$\uparrow$} & \multicolumn{1}{c}{M~$\uparrow$} & \multicolumn{1}{c}{R~$\uparrow$} & \multicolumn{1}{c}{S~$\uparrow$} \\
			\midrule
			\rvlnbert         &6.68&4.28&3.56&8.70&7.87&6.54 &3.48&2.60&1.59&5.69&5.30&4.59 \\
			\rvlnbert*        &7.13&4.59&3.87&8.94&8.18&7.30 &3.75&2.96&1.78&5.80&5.42&4.85 \\
			WebGUM$^\dagger$  &16.86&10.90&5.43&12.79&12.10&10.80&15.23&10.22&3.97&12.22&11.30&10.54 \\
			\midrule
			Ours              &\textbf{23.50}&\textbf{15.63}&\textbf{6.95}&\textbf{16.82}&\textbf{15.64}&\textbf{14.33} &\textbf{20.35}&\textbf{14.09}&\textbf{5.29}&\textbf{15.17}&\textbf{14.21}&\textbf{13.20}\\
			\bottomrule
		\end{tabular}%
	}
	\label{tab:nlg}%
\end{table*}%

\begin{table*}[t!]
	\centering
	\caption{We show the effect of $\eta$ in our WebVLN-Net on the WebVLN-v1 dataset.}
	\resizebox{0.85\linewidth}{!}
	{
		\begin{tabular}{c|cccc|cc|cccc|cc}
			\toprule
			\multirow{2}[2]{*}{$\eta$} & \multicolumn{6}{c|}{Val} & \multicolumn{6}{c}{Test} \\
			& \multicolumn{1}{c}{SR~$\uparrow$} & \multicolumn{1}{c}{OSR~$\uparrow$} & \multicolumn{1}{c}{SPL~$\uparrow$} & \multicolumn{1}{c}{TL~$\downarrow$} & \multicolumn{1}{|c}{WUPS0.9~$\uparrow$} & \multicolumn{1}{c|}{WUPS0.0~$\uparrow$} & \multicolumn{1}{c}{SR~$\uparrow$} & \multicolumn{1}{c}{OSR~$\uparrow$} & \multicolumn{1}{c}{SPL~$\uparrow$} & \multicolumn{1}{c}{TL~$\downarrow$} & \multicolumn{1}{|c}{WUPS0.9~$\uparrow$} & \multicolumn{1}{c}{WUPS0.0~$\uparrow$} \\
			\midrule
			0.1&36.93&36.93&36.93&3.97&22.90&29.99 &33.50&33.50&33.39&4.53&21.87&27.79\\
			0.2&37.16&37.16&37.16&3.52&23.15&30.20 &34.17&34.20&34.13&3.93&21.91&28.16\\
			0.5&37.24&37.24&37.15&3.64&23.25&30.36 &34.20&34.20&34.13&3.92&21.98&28.21 \\
			1  &\textbf{39.46}&\textbf{39.54}&\textbf{39.46}&3.71&\textbf{24.26}&\textbf{31.87} &\textbf{34.76}&\textbf{34.80}&\textbf{34.59}&4.34&\textbf{22.13}&\textbf{28.58}\\
			2  &37.16&37.16&37.16&3.57&21.75&28.77 &33.46&33.46&33.44&3.93&20.75&26.39\\
			10 &22.66&22.66&22.62&\textbf{3.27}&14.63&18.69 &18.84&18.92&18.78&\textbf{3.50}&13.18&16.00\\
			\bottomrule
		\end{tabular}%
	}
	\label{tab:eta}%
\end{table*}%

\begin{table*}[t!]
	\centering
	\caption{We show the effect of $\lambda$ in our WebVLN-Net on the WebVLN-v1 dataset.}
	\resizebox{0.85\linewidth}{!}
	{
		\begin{tabular}{c|cccc|cc|cccc|cc}
			\toprule
			\multirow{2}[2]{*}{$\lambda$} & \multicolumn{6}{c|}{Val} & \multicolumn{6}{c}{Test} \\
			& \multicolumn{1}{c}{SR~$\uparrow$} & \multicolumn{1}{c}{OSR~$\uparrow$} & \multicolumn{1}{c}{SPL~$\uparrow$} & \multicolumn{1}{c}{TL~$\downarrow$} & \multicolumn{1}{|c}{WUPS0.9~$\uparrow$} & \multicolumn{1}{c|}{WUPS0.0~$\uparrow$} & \multicolumn{1}{c}{SR~$\uparrow$} & \multicolumn{1}{c}{OSR~$\uparrow$} & \multicolumn{1}{c}{SPL~$\uparrow$} & \multicolumn{1}{c}{TL~$\downarrow$} & \multicolumn{1}{|c}{WUPS0.9~$\uparrow$} & \multicolumn{1}{c}{WUPS0.0~$\uparrow$} \\
			\midrule
			0.1 &37.48&37.56&37.19&\textbf{3.29}&17.70&27.93 &34.15&34.17&34.09&3.85&16.59&25.83 \\
			0.5 &38.32&38.32&38.32&3.54&22.32&29.93 &34.33&34.33&34.17&3.79&21.16&28.32 \\
			1  &\textbf{39.46}&\textbf{39.54}&\textbf{39.46}&3.71&\textbf{24.26}&\textbf{31.87} &\textbf{34.76}&\textbf{34.80}&\textbf{34.59}&4.34&\textbf{22.13}&\textbf{28.58}\\
			2   &37.16&37.16&37.16&3.51&23.47&30.33 &34.50&34.56&34.45&3.89&21.31&28.36 \\
			10  &23.38&23.38&23.25&3.46&15.99&19.66 &19.40&19.42&19.27&\textbf{3.59}&13.63&16.51 \\
			\bottomrule
		\end{tabular}%
	}
	\label{tab:lambda}%
\end{table*}%

\section{More Discussions}

\paragraph{Natural Language Generation (NLG) Evaluation}

In this part, we conduct a natural language generation (NLG) evaluation since the generated answers on our benchmark are in a free-form natural language sentence, which is easier to transit to downstream applications such as conversational agents and voice assistants.
Specifically, we consider BLEU~\cite{papineni2002bleu}, ROUGE~\cite{lin2004rouge}, METEOR~\cite{banerjee2005meteor}, CIDEr~\cite{vedantam2015cider}, and SPICE~\cite{anderson2016spice}, which are widely used on text generation tasks (\eg, image captioning) and can assess the alignment between two sentences.
As shown in Table~\ref{tab:nlg}, the proposed WebVLN-Net consistently outperforms all the baselines, including both VLN and web-related navigation methods, which further demonstrates the effectiveness of our model.

\paragraph{Impact of $\eta$}

To investigate the impact of $\eta$ in Eq.~(4), we set $\eta\in\{0.1, 0.2, 0.5, 1, 2, 10\}$.
As shown in Tables~\ref{tab:lambda}, in the case of a small $\eta$ (\eg, $\eta=0.1$), the model can only achieve suboptimal performance due to insufficient supervision from the teacher action. As we raise the value of $\eta$, the performance of our WebNet-Net reaches the peak at $\eta=1$ before declining. Taking this trade-off into account, we opt to use a weighting parameter $\eta$ of 1 for all our experiments.

\paragraph{Impact of $\lambda$} We set $\lambda\in\{0.1, 0.5, 1, 2, 10\}$ to test the effect of $\lambda$ in Eq.~(6).
From Table~\ref{tab:lambda}, when $\lambda$ is set to 1, it signifies an equal emphasis on both tasks, leading to the best overall performance. Deviating from this balanced value, either by increasing or decreasing $\lambda$, results in a reduction in performance. Thus, to strike a well-balanced performance across different scenarios, we select a consistent weighting parameter $\lambda$ of 1 for all our experiments.

\paragraph{Qualitative Results}

\begin{figure*}[htbp]
	\centering
	\includegraphics[width=0.95\linewidth]{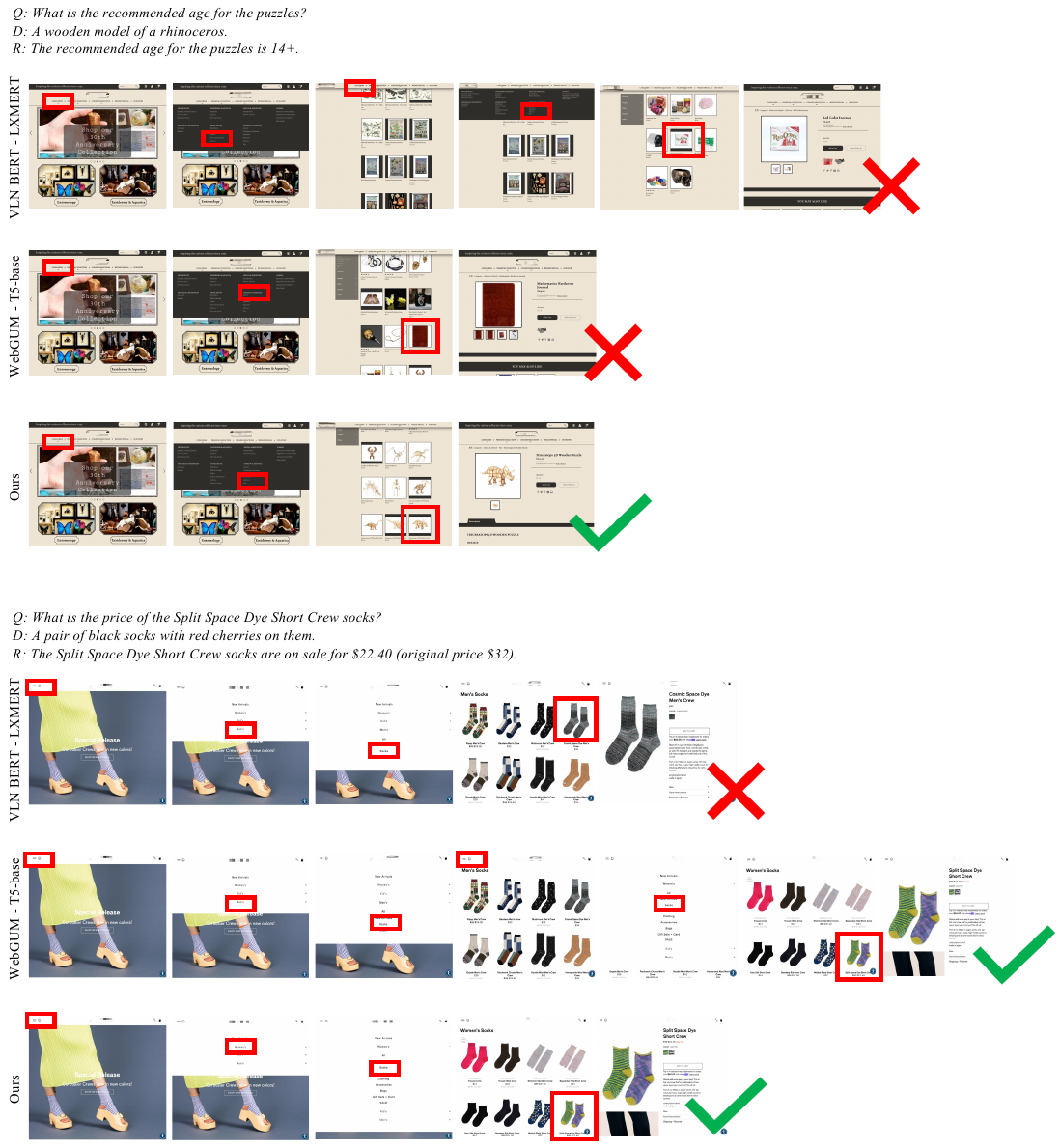}
	\caption{Visualisation of several WebVLN results, including both baseline methods and our WebVLN-Net. The red bounding box presents the action predicted. The successful navigation is indicated by the green checkmark ($\surd$), whereas navigation failure is denoted by the red cross ($\times$).}
	\label{fig:qualitative_results}
\end{figure*}

As shown in Figure~\ref{fig:qualitative_results}, we provide the visualisation of several WebVLN results obtained by baseline methods, \ie, the widely used VLN method -- \rvlnbert~and state-of-the-art (SoTA) web-related navigation method -- WebGUM, as well as our WebVLN-Net.
Compared with the baseline methods, the proposed WebVLN-Net can well reach the right target webpage in as short as possible path, which demonstrates the effectiveness of our model in both navigation and QA tasks.

\paragraph{Limitations and Future Works}

While our model demonstrates enhanced performance relative to the baseline methods, it is crucial to recognise several limitations that could be addressed in future research.
One of the limitations lies in that the constructed dataset is constrained to shopping websites due to policy considerations. In the future, we aim to augment the dataset by incorporating a wider range of website genres beyond shopping, allowing for a more comprehensive evaluation of the WebVLN task.
Furthermore, we currently restrict the agent to commence navigation only from the homepage of each website. In the future, we plan to expand the initial starting location to include any webpage within each website.

\end{document}